# Lane Change Trajectory Planning for Personalized Driving Comfort and Mobility Efficiency

Haoxuan Dong
*Department of Mechanical Engineering*
*National University of Singapore*
Singapore 117575, Singapore
donghaox@foxmail.com

Dongjun Li
*Department of Mechanical Engineering*
*National University of Singapore*
Singapore 117575, Singapore
dongjun.li@u.nus.edu

Ziyou Song
*Department of Electrical Engineering and Computer Science*
*University of Michigan*
Ann Arbor, MI 48109, USA
ziyou@umich.edu

***Abstract*—Lane changing entails simultaneous longitudinal and lateral motions that affect driving comfort and mobility efficiency. Because these motions are tightly coupled and subject to substantial inter-vehicle variability, trajectory planning for lane-change maneuvers is characterized by a highly personalized nature. This study proposes a neural network-driven planner that integrates a third-order polynomial trajectory generator with a learning module that infers optimal trajectory parameters across diverse driving conditions. Using a shared backbone with dual heads, one head ensures all-condition operational guarantees, while the other captures driver-specific preferences for comfort or mobility efficiency. A head-gated switching mechanism, realized through a statistical gate based on error-winner logistic regression, adaptively selects the appropriate head under varying driving conditions, which enables context-aware lane-change trajectory planning. Representative cases and Monte Carlo simulations show that the proposed planner achieves personalized comfort and mobility during lane changes, while the baseline ensures feasible trajectories under driving conditions where personalized data are insufficient or inaccessible.**

## I. INTRODUCTION

Lane changing is one of the most frequent maneuvers in daily driving, typically executed when a vehicle needs to overtake slower traffic or avoid obstacles [1]. Unlike longitudinal motion control, however, it requires the simultaneous generation of longitudinal and lateral trajectories and is strongly influenced by surrounding vehicles in both the current and target lanes [2]. This coupling effect not only increases the complexity of lane-change trajectory planning but also amplifies the associated safety risks. Moreover, because lane-change maneuvers are usually performed within short time horizons, trajectory planning faces the significant challenge of ensuring safety while concurrently satisfying comfort and mobility requirements in a well-coordinated manner [3].

Traditional studies on lane-change trajectory planning employed model-based formulations using polynomial parameterizations, such as third-order [4] and seventh-order [5] curves, to generate smooth trajectories. These methods typically impose lateral acceleration constraints; however, their reliance on fixed polynomial families and a single time-oriented objective limits adaptability to diverse traffic interactions and overlooks the essential trade-offs among safety, comfort, and mobility. This motivates the adoption of optimization-based lane-change planning. Nguyen *et al.* [6] combined the Pontryagin maximum principle with a time-to-collision metric to generate near-optimal lane-change trajectories that enhance comfort while preserving safe distances. Lin *et al.* [7] proposed a hierarchical planner that decouples path and speed, initialized via sequential quadratic programming and refined with curvature smoothing. Li *et al.* [8] maximized the joint benefit of the ego and surrounding vehicles and obtained the optimal lane-change trajectory using a genetic algorithm.

Recently, learning-based methods have emerged as a complementary direction to optimization-based formulations for lane-change trajectory planning. Leveraging demonstrations and interaction data, imitation learning enables the distillation of expert behavior, while deep reinforcement learning facilitates online trajectory planning in interactive traffic. Meng *et al.* [9] developed a multi-task deep learning model for lane-change trajectory prediction to enhance driving safety. Sun *et al.* [10] introduced a segmented-update method that generates lane-change trajectories by learning cost functions with maximum entropy inverse reinforcement learning and refining them through online optimization. Jing *et al.* [11] proposed a deep reinforcement learning-based method for lane-change trajectory planning, employing a long short-term memory-based twin delayed deep deterministic policy gradient algorithm to enhance efficiency and reduce fuel use.

Despite these advances, most existing methods [6-11] still rely on fixed objective weights, which limits their ability to adapt to individual drivers' preferences regarding the trade-off between comfort and mobility. Some personalized lane-change studies, such as Nie *et al.* [3], attempt to address this by categorizing drivers into discrete classes (e.g., cautious, moderate, aggressive) and constraining parameter ranges accordingly, with online classifiers trained on driver data. However, such coarse classification inevitably oversimplifies the diversity of driving behaviors and restricts personalization to a few predefined categories, failing to capture the nuanced variability in individual driving styles.

We propose a novel lane-change trajectory planner that generates trajectories while conveniently accommodating personalized requirements for driving comfort and mobility efficiency. The contributions are threefold. First, we integrate a third-order parametric polynomial with a multilayer perceptron (MLP) to capture features that characterize personalized driving

This work was supported in part by the A-Star Young Individual Research Grants (YIRG), Singapore, under Grant M22K3c0092.

preferences and enable fast and accurate computation of optimal lane-change trajectories across diverse driving conditions. Second, we propose a shared-backbone dual-head architecture in which one head ensures baseline feasibility while the other captures driver-specific preferences, thus balancing universal requirements with personalized adaptation. Third, we design a head-gated statistical switching mechanism, which combines error-weighted logistic regression, to adaptively select baseline or personalized heads, according to driving conditions.

The remainder of this paper is organized as follows. Section II formulates vehicle kinematics and trajectory modeling. Section III presents the problem, framework, and theoretical foundations of the proposed lane-change trajectory planner. The performance of the proposed planner is comprehensively evaluated in Section IV. Finally, Section V concludes the paper.

## II. Vehicle Kinematics and Trajectory Modelling

### A. Vehicle Kinematics

This study focuses on routine lane-change scenarios, without explicitly considering tire slip or steering dynamics. Under these conditions, the maneuver can be represented as a smooth geometric trajectory in the spatial domain. The kinematic relations of distance, speed, and acceleration, accounting for longitudinal and lateral motion, are given in (1)-(3).

$$s = \int_0^{s_x} \sqrt{1 + \left(\frac{\mathrm{d}s_y}{\mathrm{d}s_x}\right)^2}\, \mathrm{d}s_x \tag{1}$$

$$v = \sqrt{v_x^2 + v_y^2} \tag{2}$$

$$a = \sqrt{a_x^2 + a_y^2} \tag{3}$$

with

$$v_x = \frac{\mathrm{d}s_x}{\mathrm{d}t}, v_y = \frac{\mathrm{d}s_y}{\mathrm{d}t}, a_x = \frac{\mathrm{d}v_x}{\mathrm{d}t}, a_y = \frac{\mathrm{d}v_y}{\mathrm{d}t}$$

where $s$ is the arc length along the trajectory, $s_y$ and $s_x$ are the lateral and longitudinal displacements, respectively. $v$ is the speed along the vehicle heading, $v_y$ and $v_x$ are the lateral and longitudinal speeds, respectively. $a$ is the acceleration along the vehicle heading, $a_y$ and $a_x$ are the lateral and longitudinal accelerations, respectively.

### B. Lane-change Trajectory Generation

We use a third-order polynomial to generate lane-change trajectory [4, 12]. This polynomial provides a closed-form representation with continuous second-order derivatives, thereby ensuring a smooth trajectory in the spatial domain:

$$s_y(s_x) = \gamma_0 + \gamma_1 s_x + \gamma_2 s_x^2 + \gamma_3 s_x^3 \tag{4}$$

where $\gamma_*$ are the polynomial coefficients. Without loss of generality, the vehicle at the start of the lane-change maneuver is assumed to be aligned horizontally, with its longitudinal and lateral positions as well as yaw angle set to zero. At the end of the maneuver, the longitudinal and lateral positions are specified as $\mathcal{S}_x$ and $\mathcal{S}_y$, respectively, while the heading angle remains horizontal. By substituting these initial and terminal conditions into (4), $\gamma_*$ are obtained as follows:

$$\gamma_0 = 0, \gamma_1 = 0, \gamma_2 = \frac{3\mathcal{S}_y}{\mathcal{S}_x^2}, \gamma_3 = \frac{-2\mathcal{S}_y}{\mathcal{S}_x^3} \tag{5}$$

Substituting (5) into (4) yields:

$$s_y(s_x) = \frac{3\mathcal{S}_y}{\mathcal{S}_x^2} s_x^2 - \frac{2\mathcal{S}_y}{\mathcal{S}_x^3} s_x^3 \tag{6}$$

Given the relatively small variations in $v_x$ compared with lateral dynamics of lane-change maneuvers, a constant $v_x$ is assumed, which constitutes a widely adopted simplification in the lane-change study [13]. According to (6), $a_y$ is obtained by the chain rule as the second derivative of $s_y$ with respect to $s_x$ multiplied by $v_x^2$. This yields:

$$a_y(s_x) = v_x^2 \frac{\mathrm{d}^2 s_y}{\mathrm{d}s_x^2} = \frac{6\mathcal{S}_y v_x^2}{\mathcal{S}_x^3}(\mathcal{S}_x - 2s_x) \tag{7}$$

It follows that $a_y(s_x)$ varies linearly with $s_x$ and reaches symmetric extrema at the start and end of the maneuver. Then the peak lateral accelerations $|a_y|_{\max}$ are

$$|a_y|_{max} = \frac{6\mathcal{S}_y}{\mathcal{S}_x^2} \max\{\mathcal{V}_s^2, \mathcal{V}_f^2\} \tag{8}$$

where $\mathcal{V}_s$ and $\mathcal{V}_f$ are the speed at the start and end of the maneuver, respectively. Since $a_x = 0$ under the constant $v_x$ assumption, substituting (7) into (3) gives $a(s_x) = |a_y(s_x)|$.

The instantaneous time increment $dt$ is computed from the arc-length element $\mathrm{d}s$ and the speed $v(s_x)$. Specifically,

$$\mathrm{d}t = \frac{\mathrm{d}s}{v(s_x)} \tag{9}$$

with

$$v(s_x) = \sqrt{v_x^2(s_x) + v_y^2(s_x)}, \mathrm{d}s = \sqrt{1 + \left(\frac{\mathrm{d}s_y}{\mathrm{d}s_x}\right)^2}\, \mathrm{d}s_x$$

Accordingly, the lane-change time $t$ as a function of $s_x$ is

$$t(s_x) = \int_0^{s_x} \frac{\sqrt{1 + \left(\frac{\mathrm{d}s_y}{\mathrm{d}\xi}\right)^2}}{\sqrt{v_x^2(\xi) + v_y^2(\xi)}}\, \mathrm{d}\xi \tag{10}$$

where $\xi$ is the integration variable for longitudinal displacement. Then, the total duration $T$ is obtained by setting $s_x = \mathcal{S}_x$:

$$T = t(\mathcal{S}_x) \tag{11}$$

## III. Lane-change Trajectory Planning

### A. Problem Formulation

In lane-change maneuvers, $\mathcal{S}_y$ is usually determined by the lane width, while $\mathcal{V}_s$ is the vehicle speed at the onset of the maneuver determines the lane change. Based on the assumption of constant $v_x$ and the fact that $v_y$ is zero at the beginning and end of the lane change,, the final speed can be approximated as $\mathcal{V}_f = \mathcal{V}_s$. Consequently, the choice of $\mathcal{S}_x$ is critical, as they govern the trajectory shape and driving time, which in turn affect the vehicle driving safety, comfort, and mobility. In general, safety is a prerequisite for lane-change maneuvers. However, balancing comfort and mobility is challenging, as these two factors are interrelated [14, 15]. Moreover, the desired levels of comfort and mobility vary among drivers, reflecting individualized preferences [16].

When a lane change is required, the feasible ranges of $\mathcal{S}_x \in [\mathcal{S}_{x,min}, \mathcal{S}_{x,max}]$ can be determined from the states of the ego vehicle and surrounding traffic. Here, $\mathcal{S}_{x,min}$ and $\mathcal{S}_{x,max}$ are the lower and upper bounds of $\mathcal{S}_x$, respectively. Considering the maximum lateral acceleration limit for safety, trajectories that violate this limit are excluded in advance. The remaining trajectories in the feasible range then define a candidate set $\boldsymbol{C} = \{\mathcal{S}_{x,k}\}_{k=1}^{N}$, where each $\mathcal{S}_{x,k}$ corresponds to a distinct trajectory characterized by different lateral acceleration and lane-change time. However, exhaustive search within $\boldsymbol{C}$ is computationally expensive and time-consuming.

## B. Overall Framework

To address this limitation, we train an MLP surrogate on selected data to efficiently predict the optimal longitudinal position $\mathcal{S}_x^*$ for different driving conditions. Furthermore, a shared-backbone dual-head architecture is incorporated to extract distinct lane-change knowledge from different datasets, while a head-gated statistical switching mechanism is employed to facilitate the selection between heads, thereby enabling personalized adaptation to individual driving preferences. The overall framework of the proposed planner is shown in Fig. 1.

## C. Dual-head MLP with Shared-backbone

We adopt a shared-backbone, dual-head MLP to learn the mapping from driving conditions to the optimal longitudinal distance. Let the input be $x = [\mathcal{V}_s, \mathcal{S}_{x,min}, \mathcal{S}_{x,max}]^{\mathrm{T}} \in \mathbb{R}^3$. Two regression heads are trained: a baseline (B) head for the baseline dataset and a personalized (P) head for the comfort or mobility priority dataset. Formally, the model implements

$$f_{\theta_B}^{(B)}: \mathbb{R}^3 \to \mathbb{R},\, f_{\theta_P}^{(P)}: \mathbb{R}^3 \to \mathbb{R},\, x \mapsto \hat{\mathcal{S}}_x^{*(B)}, \hat{\mathcal{S}}_x^{*(P)} \tag{12}$$

where $\theta_B$ and $\theta_P$ are the parameters of the two heads together with the shared backbone. $\hat{\mathcal{S}}_x^{*(B)}$ and $\hat{\mathcal{S}}_x^{*(P)}$ are the predicted optimal longitudinal distance of the two heads in the original scale.

Inputs are standardized componentwise using robust statistics computed on the training split, then the robustly standardized input vector $\tilde{x}$ is

$$\tilde{x} = \frac{x - \mathrm{median}(x)}{\mathrm{IQR}_{[5,95]}(x)} \tag{13}$$

where $\mathrm{IQR}_{[5,95]}(\cdot)$ is the inter-quantile range between the 5th and 95th percentiles and $\mathrm{median}(\cdot)$ is componentwise. Targets are transformed by the Yeo-Johnson power transform fitted separately for the two domains, $\mathcal{T}_{YJ}^{(B)}$ for the baseline and $\mathcal{T}_{YJ}^{(P)}$ for personalization. Predictions are mapped back by the corresponding inverse transforms at evaluation.

The forward propagation through the shared backbone with $K$ hidden layers is

$$h_0 = \tilde{x} \in \mathbb{R}^3 \tag{14}$$

$$h_k = \psi(W_k h_{k-1} + b_k) \odot m_k,\, k = 1, 2, \dots, K \tag{15}$$

$$\hat{y}^{(B)} = w_B^{\mathrm{T}} h_K + b_B,\, \hat{y}^{(P)} = w_P^{\mathrm{T}} h_K + b_P \tag{16}$$

where $h_k$ is the hidden representation at layer $k$, $W_k$ and $b_k$ are the weight matrix and bias vector of layer $k$, respectively. $\psi(\cdot) = \mathrm{ReLU}(\cdot)$ is the activation, $m_k \sim \mathrm{Bernoulli}(1 - p_d)$ is the dropout mask with dropout rate $p_d$. $w_{(\cdot)}$ and $b_{(\cdot)}$ are the weights and bias of the output layer, respectively. $\hat{y}^{(\cdot)}$ is the transformed-scale prediction. The two heads $(w_B, b_B)$ and $(w_P, b_P)$ output predictions in the transformed spaces.

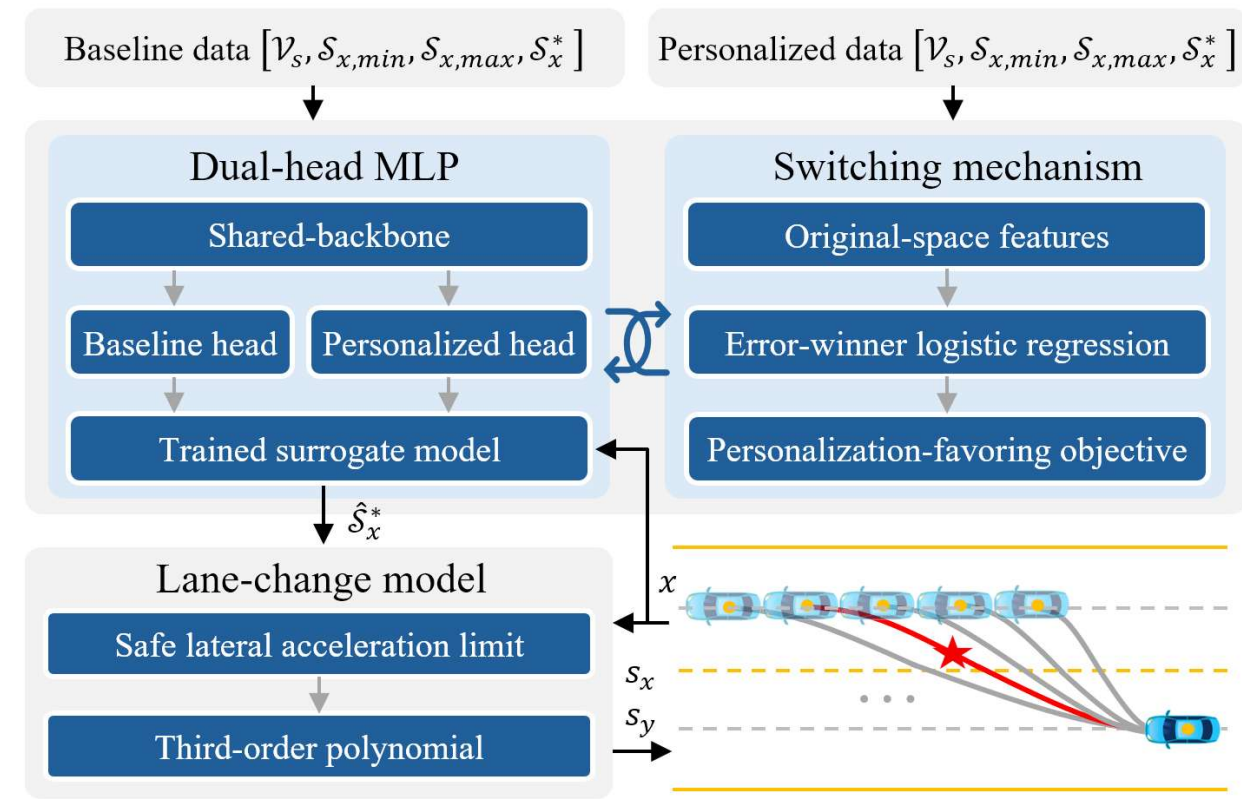

**Fig. 1.** Overall framework of the proposed lane-change trajectory planner.

For a training batch $\{(\tilde{x}_i, \tilde{y}_i)\}_{i=1}^{N}$ in a given stage, the network is optimized with Huber loss (threshold $\delta > 0$):

$$\mathcal{L}_\delta(\theta) = \frac{1}{N} \sum_{i=1}^{N} \ell_\delta(\hat{y}_i - \tilde{y}_i) \tag{17}$$

$$\ell_\delta(u) = \begin{cases} \frac{1}{2} u^2 & |u| \le \delta \\ \delta |u| - \frac{1}{2} \delta^2 & |u| > \delta \end{cases} \tag{18}$$

which is quadratic near zero and linear for large residuals. Predictions in the original target space are

$$\hat{\mathcal{S}}_x^{*(B)} = \left(\mathcal{T}_{YJ}^{(B)}\right)^{-1}(\hat{y}^{(B)}),\, \hat{\mathcal{S}}_x^{*(P)} = \left(\mathcal{T}_{YJ}^{(P)}\right)^{-1}(\hat{y}^{(P)}) \tag{19}$$

We adopt AdamW with weight decay, gradient clipping at 1.0, a ReduceLROnPlateau scheduler on validation loss, and early stopping. Training has two stages: first, the backbone and baseline head are trained to convergence; second, the personalized head is trained with the backbone frozen.

## D. Head-gated Statistical Switching Mechanism

The two-head network is combined by a statistical gate, implemented via error-winner logistic regression, that decides for each input which head to trust. The gate forms the original-space feature vector

$$u = [\mathcal{V}_s, \mathrm{width}]^{\mathrm{T}},\, \mathrm{width} = \mathcal{S}_{x,max} - \mathcal{S}_{x,min} \tag{20}$$

On validation data, the baseline squared error $e_B(u)$ and comfort-priority squared error $e_P(u)$ are defined as:

$$e_B(u) = (\hat{\mathcal{S}}_x^{*(B)} - \mathcal{S}_x^*)^2,\ e_P(u) = (\hat{\mathcal{S}}_x^{*(P)} - \mathcal{S}_x^*)^2 \tag{21}$$

The error-winner label $z(u)$, denoting the head that achieves lower squared error at input $u$, is then given by

$$z(u) = \mathbb{I}(e_P(u) < e_B(u)) \tag{22}$$

where $\mathbb{I}(\cdot)$ is the indicator function, which returns 1 if its logical argument is true and 0 otherwise. $z(u)$ serves as supervision for learning a gating model, which predicts the probability that the personalized head outperforms the baseline head given input $u$.

A logistic regression trained on $(u|z(u))$ yields the gating probability:

$$\mathcal{P}_P(u) = \Pr(\mathrm{z} - 1|u) \tag{23}$$

At inference, $\mathcal{P}_P(u)$ is thresholded to decide which head's output to adopt, as expressed in (24).

$$\hat{\mathcal{S}}_x^* = \begin{cases} \hat{\mathcal{S}}_x^{*(P)} & \mathcal{P}_P(u) > \tau \\ \hat{\mathcal{S}}_x^{*(B)} & \mathcal{P}_P(u) \le \tau \end{cases} \tag{24}$$

where the threshold $\tau$ is selected on the validation set by grid search. The optimal value is determined by minimizing a proxy objective that encourages personalization gains while safeguarding baseline performance:

$$J(\tau) = r_P(\tau) + \varpi \max\left(0, r_B(\tau) - r_B^{\mathrm{S1}}(\tau) - \Delta\right) \tag{25}$$

where $r_P(\tau)$ and $r_B(\tau)$ are the root mean squared error of the gated predictor on the personalized and baseline sets at threshold $\tau$, respectively. $r_B^{\mathrm{S1}}$ is the Stage 1 baseline root mean squared error, $\varpi > 0$ balances personalized improvement against baseline protection, and $\Delta \ge 0$ specifies the tolerated slack for baseline root mean squared error increase. This design retains Stage 1 capability under baseline-like conditions while adapting to personalized priorities, without requiring further backpropagation after Stage 2.

In applications, the trained surrogate model takes $\mathcal{V}_s$ together with $\mathcal{S}_{x,min}$ and $\mathcal{S}_{x,max}$ determined by driving conditions as inputs, and outputs $\hat{\mathcal{S}}_x^*$. These inputs and the predicted output are then incorporated into the third-order polynomial to generate the corresponding lane-change trajectory.

# IV. Simulation and Results

## A. Simulation Setup

The surrogate with three hidden layers [64, 32, 16], trained using AdamW with an initial learning rate of $1\times10^{-3}$, weight decay of $1\times10^{-4}$, batch size 32, dropout rate of 0, and early stopping with a patience of 100. The backbone learning rate is $5\times10^{-4}$. For the error-winner logistic regression, the decision threshold $\tau$ is selected on a grid from 0.2 to 0.8 with a step size of 0.02, and the proxy objective is evaluated with $\varpi = 0.8$ and $\Delta = 12\text{m}$. The optimal threshold obtained on the validation set is 0.7. Hyperparameters are manually tuned to achieve the best performance. All datasets are partitioned into training, validation, and test sets with a ratio of 80%, 10%, and 10%, respectively. The simulation is performed in Python on a workstation equipped with an Intel i7-13700KF CPU, 64GB RAM, and an NVIDIA RTX 4080 GPU running Windows 10.

## B. Training Datasets

We generated three datasets to evaluate the performance of the proposed planner, namely baseline, comfort-priority, and mobility-priority, using third-order polynomial trajectories under diverse driving conditions. Their only difference lies in the maximum lateral-acceleration constraint: 0.12g for baseline, 0.03g for comfort, and 0.30g for mobility, following Ksander *et al*. [17]. These limits approximately correspond to "So-so," "Excellent," and "Terrible" on their verbal comfort scale.

The baseline dataset represents commonly acceptable conditions and is generated by uniform gridding, with $\mathcal{V}_s$ ranging from 20 to 120 km/h in 5 km/h steps and $\left[\mathcal{S}_{x,max}, \mathcal{S}_{x,min}\right]$ from 5 to 250 m in 5 m steps. In contrast, the comfort and mobility datasets capture individualized preferences that emphasize passenger comfort or rapid lane changes, respectively. They are generated under their respective acceleration limits using Latin hypercube sampling with distributional biasing (Figs. 2b-2c). The comfort dataset focuses on medium-to-high speeds and moderate-to-long distances, whereas the mobility dataset is more uniformly distributed with stronger coverage at medium-to-low speeds and short-to-moderate distances.

## C. Benchmarks

To evaluate the advantages of the proposed lane-change planner in achieving personalized performance, three benchmarks are introduced. Benchmarks 1, 2, and 3 correspond to conventional MLP-based planners trained solely on the baseline, comfort-priority, and mobility-priority datasets, respectively. For fairness, the MLP configurations used in these benchmarks are identical to those adopted in the designed dual-head MLP. The proposed planner trains its baseline head on the baseline dataset, while the personalized head is trained on the comfort-priority and mobility-priority datasets, resulting in two personalized variants denoted as Planner-Pcomf and Planner-Pmobi. This design enables the planner to retain the general applicability of the baseline model while adapting to individualized preferences for comfort and mobility efficiency.

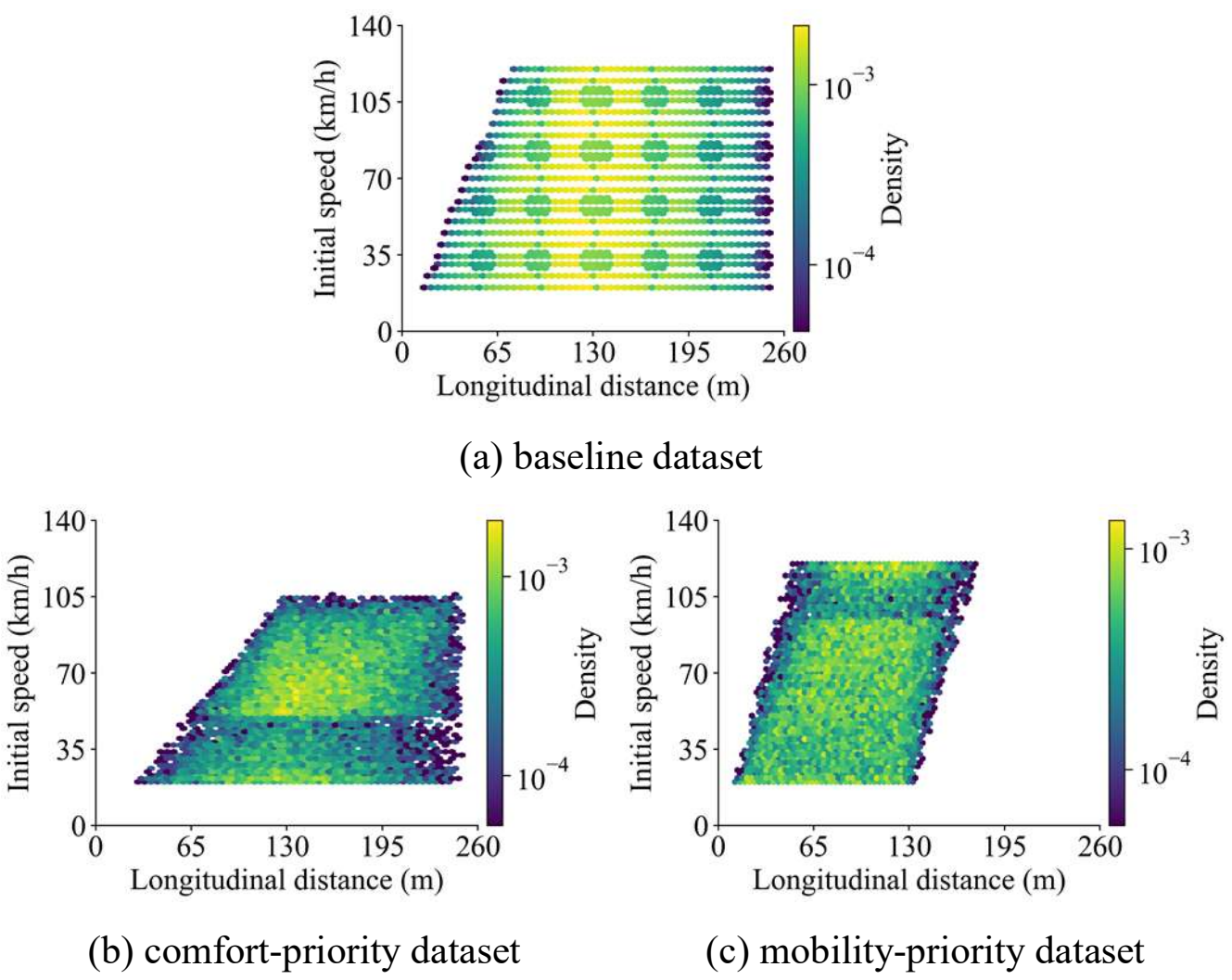

**Fig. 2.** The distribution of initial speeds and longitudinal distance across baseline, comfort-priority, and mobility-priority datasets.

*D. Results*

*1) Dual-head MLP Training Results:* The results for one training instance (i.e., the dual-head MLP combining baseline and comfort-priority heads) are presented in Fig. 3, including the prediction performance on the validation set and the decision boundary of the head-gated statistical switching mechanism.

In Figs. 3(a) and 3(b), the validation data of the baseline head is tightly clustered around the prediction line, indicating high accuracy, whereas that of the comfort-priority head is more scattered but still aligned with the prediction line, reflecting its specialization for comfort conditions. In Fig. 3(c), the 0.7 probability contour marks the division of responsibility between the two heads: the comfort-priority head governs at higher speeds and longer distances, whereas the baseline head dominates at lower speeds and shorter distances. This boundary is consistent with the training distribution shown in Fig. 2(b).

These results indicate that the dual-head MLP with a shared backbone successfully captures the distinct characteristics of both baseline and comfort-priority datasets. The statistical gating mechanism further exploits their distributional differences to select the most suitable head, with the baseline head providing fundamental guarantees and the personalized head leveraging the unique features of individualized data.

*2) Representative Case Study:* A representative lane-change case is selected with an initial speed of 97.40km/h and a longitudinal distance ranging from 67.66m to 239.04m. The vehicle position and lateral acceleration are shown in Fig. 4.

Under different lane-change trajectory plan strategies, as displayed in Fig. 4, a higher maximum lateral acceleration during lane change corresponds to a shorter maneuver duration, while trajectory smoothness (comfort) is inversely related to the attainable maximum lateral acceleration. Under Benchmark 2, the vehicle achieves the lowest maximum lateral acceleration, resulting in the longest lane-change time (poorest mobility) but the smoothest longitudinal trajectory (highest comfort). In contrast, Benchmark 3 allows the highest maximum lateral acceleration, yielding the shortest lane-change time (best mobility) but the most abrupt longitudinal trajectory (poorest comfort). Benchmark 1 lies between these extremes, leaning toward comfort. These results confirm that each benchmark meets the performance targets of its dataset.

The lateral position trajectory of Planner-Pcomf falls between Benchmark 1 and Benchmark 2, closer to the latter, indicating that the planner has learned comfort-priority driving features on top of the baseline. Conversely, the trajectory of Planner-Pmobi lies between Benchmark 1 and Benchmark 3, closer to Benchmark 3, showing that it has captured mobility-priority driving features while retaining the baseline foundation. These results demonstrate that the proposed planner can effectively accommodate individualized comfort or mobility requirements during lane-change maneuvers.

*3) Stochastic Simulation:* To evaluate the generalizability and reliability of the proposed planner, 10000 Monte Carlo simulations are performed with randomized initial speeds and longitudinal distance ranges. The resulting distributions of lane-change duration and maximum lateral acceleration under different plan strategies are illustrated in Fig. 5.

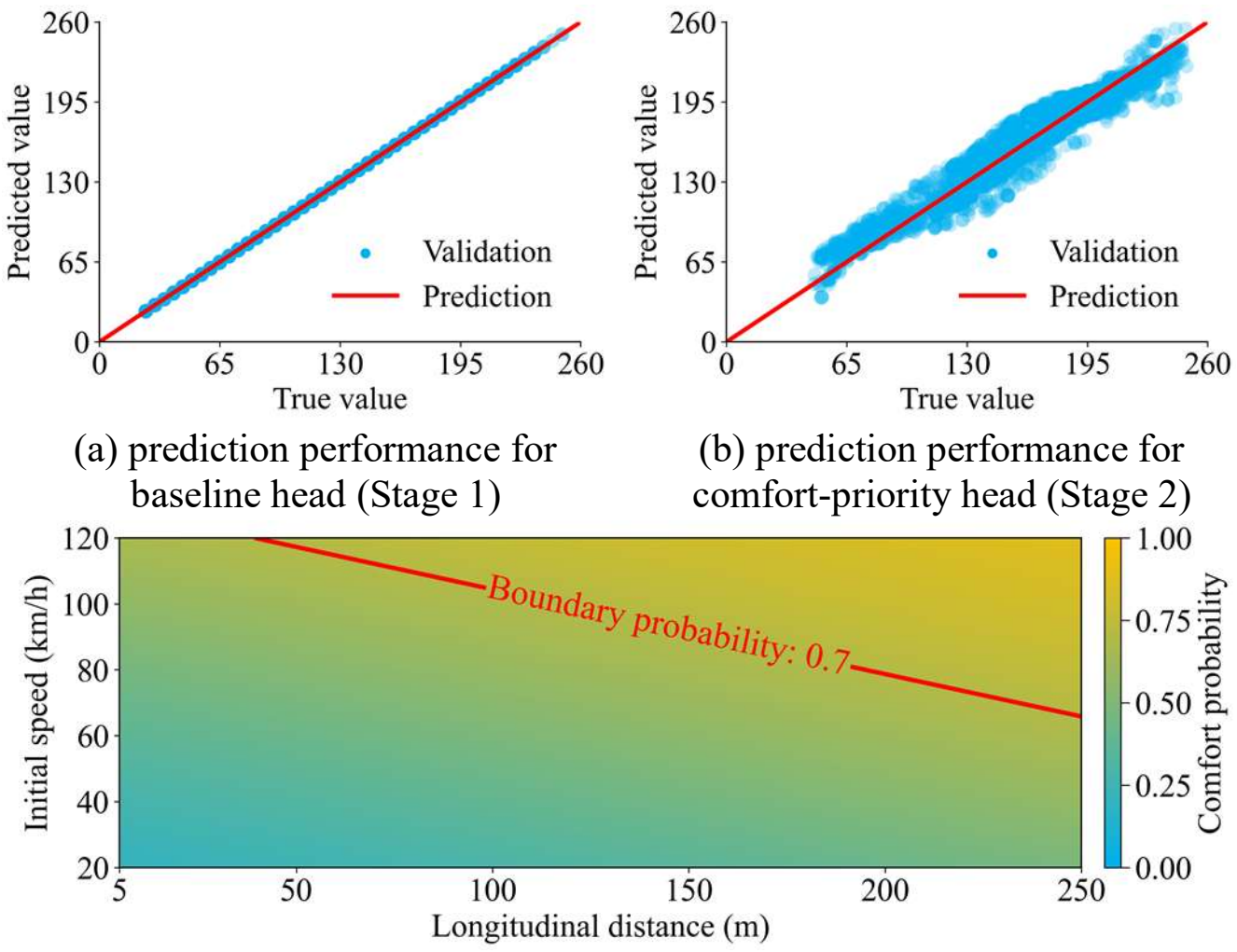

(a) prediction performance for baseline head (Stage 1)

(b) prediction performance for comfort-priority head (Stage 2)

(c) decision boundary of the head-gated statistical switching mechanism

**Fig. 3.** Results of one training instance.

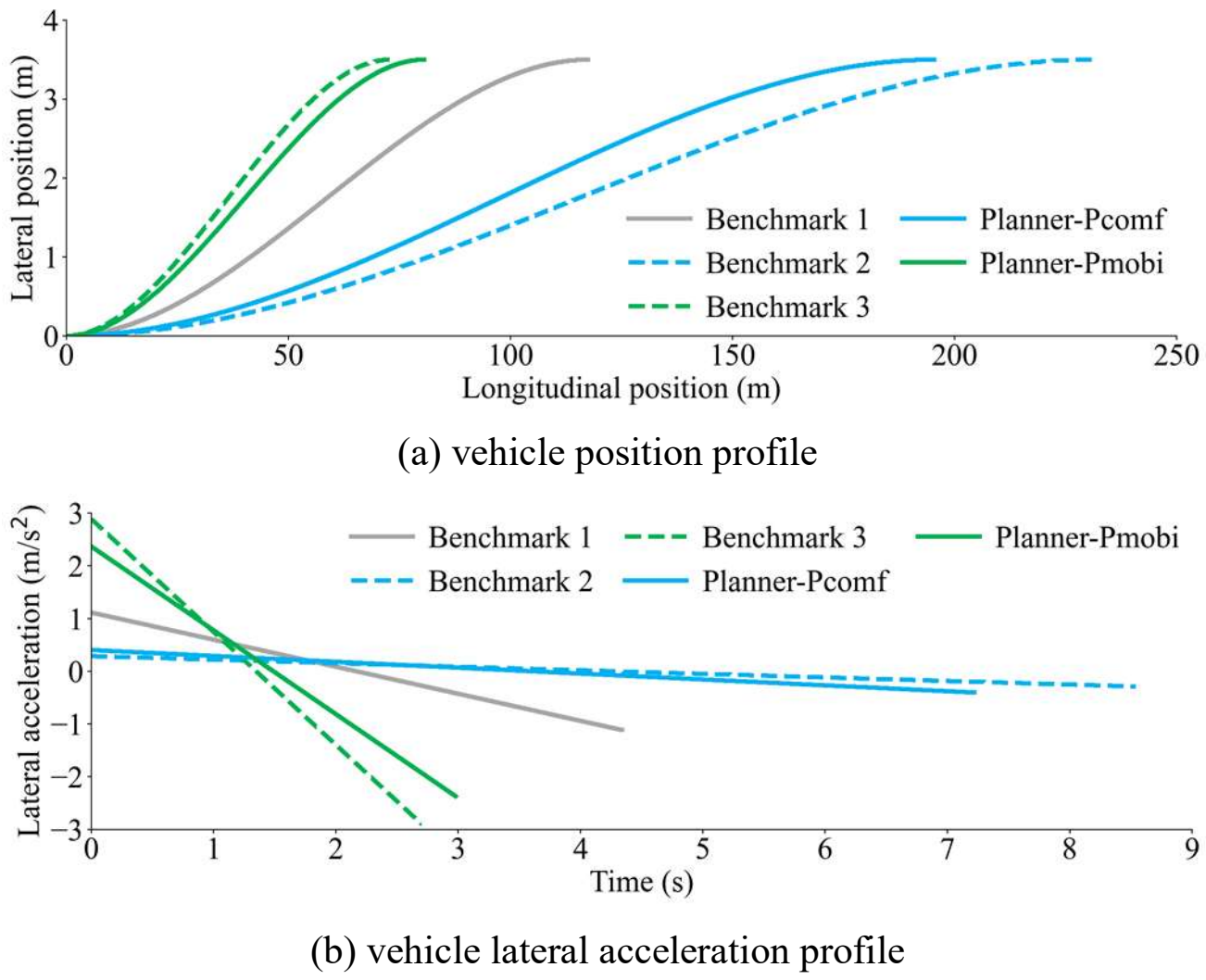

(a) vehicle position profile

(b) vehicle lateral acceleration profile

**Fig. 4.** Results of representative lane-change case.

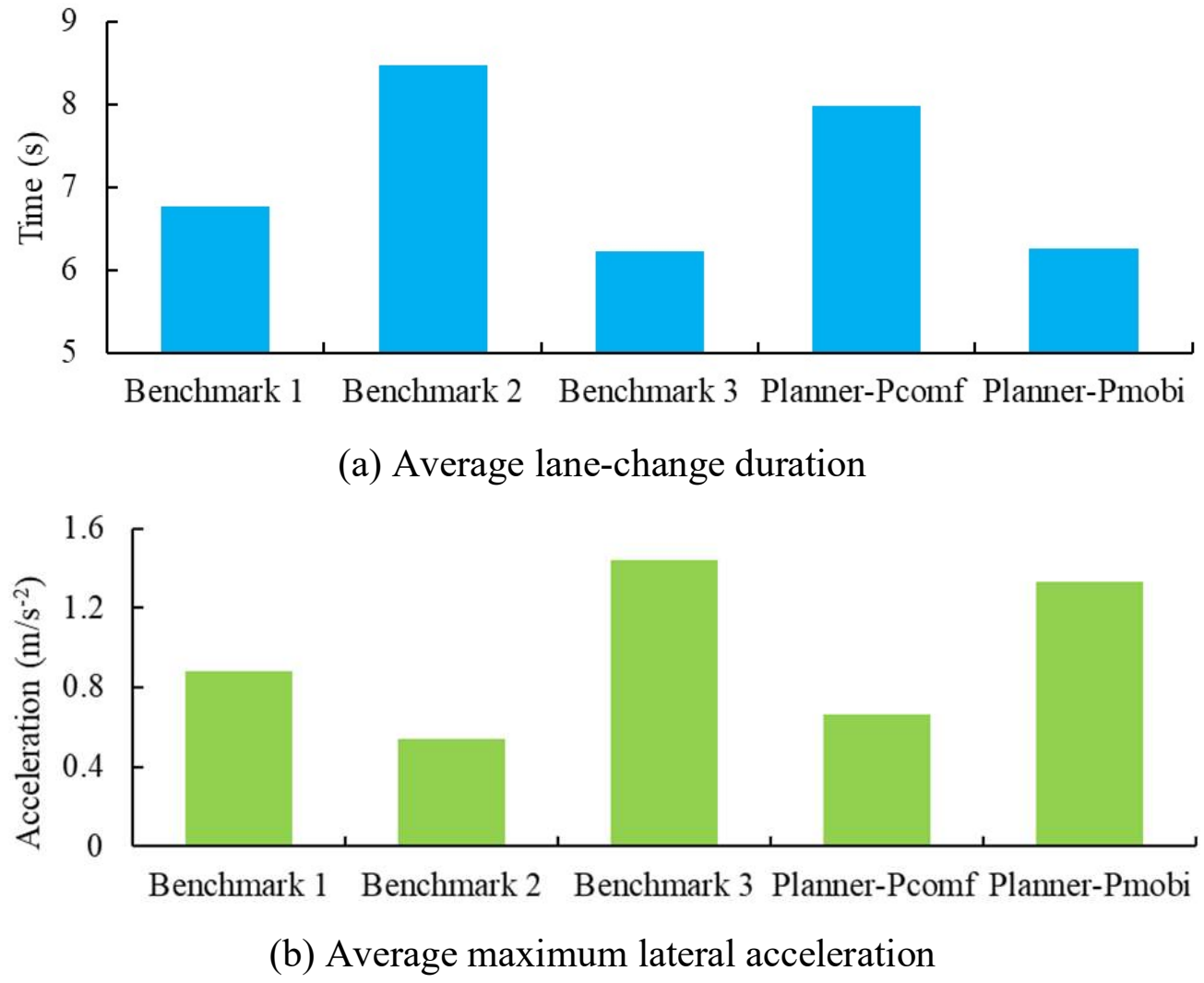

(a) Average lane-change duration

(b) Average maximum lateral acceleration

**Fig. 5.** Results of Monte Carlo simulations.

As shown in Fig. 5, the characteristics of average lane-change duration and average maximum lateral acceleration are consistent with the conclusions drawn from the representative case study. Benchmark 2 achieves the lowest average maximum lateral acceleration ($0.54\text{m/s}^2$) and thus the longest average lane-change time (8.47s). In contrast, Benchmark 3 yielded the highest average maximum lateral acceleration ($1.44\text{m/s}^2$) and the shortest average duration (6.23s), while Benchmark 1 falls between the two. The personalized planners further illustrate this trade-off: Planner-Pcomf reduces the average maximum lateral acceleration to $0.66\text{m/s}^2$ relative to Benchmark 1 ($0.88\text{m/s}^2$), but remains higher than Benchmark 2, whereas Planner-Pmobi increases it to $1.33\text{m/s}^2$ compared with Benchmark 1, but still below Benchmark 3. These results indicate that the proposed planners exhibit no performance deviation under randomized scenarios and achieve the expected outcomes.

### *E. Discussions*

Taken together with the representative case and Monte Carlo simulation results, these findings confirm that the proposed planners can effectively adapt the baseline behavior to incorporate individualized preferences for either comfort or mobility. This property brings notable convenience to trajectory planning and control design: a baseline lane-change planner can first be pre-trained with theoretical data that covers a wide range of driving conditions, and later be personalized after deployment by incorporating actual lane-change data collected from drivers. Such a design can yield trajectory planners that better align with user preferences, which is a key focus of autonomous driving or driving assistance systems [18]. Moreover, our approach alleviates the limitations caused by the long-tail effect in driving data collection [19], where the rareness condition may otherwise bias training based solely on driver data, since the baseline ensures robust performance even in these infrequent conditions.

## V. Conclusion

This study proposed a lane-change trajectory planner capable of achieving personalized driving comfort and mobility efficiency. The planner integrates a neural network with a third-order polynomial, ensuring lane-change trajectory planning is efficient. By employing a shared-backbone dual-head MLP, the framework simultaneously learns baseline features and personalized lane-change characteristics. In addition, a head-gated statistical switching mechanism with error-winner logistic regression is incorporated to enable adaptation across diverse traffic conditions. Simulation results show that the proposed planner leverages personalized driving data to capture lane-change characteristics, with the baseline head ensuring fundamental guarantees. It generates individualized trajectories that adapt to comfort or mobility preferences defined in the training datasets. This provides a new perspective for designing lane-change planners tailored to individual vehicles in future autonomous and assisted driving. Future work will focus on integrating the proposed planner into advanced driver assistance or autonomous driving pipelines, enabling personalized lane-change planning with closed-loop control.